\documentclass{article}
\usepackage{fancyhdr}
\usepackage{spconf,amsmath,graphicx}
\usepackage{subcaption}

\usepackage{bbm}

\title{PhyOT: Physics-informed object tracking in surveillance cameras}
  \name{Kawisorn Kamtue$^{\star}$ \qquad Jose M.F. Moura$^{\star}$ \qquad Orathai Sangpetch$^{\dagger}$ \qquad Paulo Garcia$^{\S}$ \sthanks{This work is supported by Thaibev in colloboration with CMKL university.}}
  \address{$^{\star}$ Carnegie Mellon University \\
      $^{\dagger}$ CMKL University \\
      $^{\S}$ Chulalongkorn University} 

\begin{document}
\ninept

\markboth{Preprint version: published version at IEEE ICASSP 2024.}%
{Kamtue \MakeLowercase{\textit{et al.}}: PhyOT}

\thispagestyle{fancy}
\fancyhead{} 
\fancyhead[RO,LE]{\tiny Preprint version: published version accepted at IEEE ICASSP 2024 \copyright 2023 IEEE. Personal use of this material is permitted.
  Permission from IEEE must be obtained for all other uses, in any current or future 
  media, including reprinting/republishing this material for advertising or promotional 
  purposes, creating new collective works, for resale or redistribution to servers or 
  lists, or reuse of any copyrighted component of this work in other works.}

\maketitle

\begin{abstract}

While deep learning has been very successful in computer vision, real world operating conditions such as lighting variation, background clutter, or occlusion hinder its accuracy across several tasks. Prior work has shown that hybrid models---combining neural networks and heuristics/algorithms---can outperform vanilla deep learning for several computer vision tasks, such as classification or tracking. 
\par We consider the case of object tracking, and evaluate a hybrid model (PhyOT) that conceptualizes deep neural networks as ``sensors'' in a Kalman filter setup, where prior knowledge, in the form of Newtonian laws of motion, is used to fuse sensor observations and to perform improved estimations. Our experiments combine three neural networks, performing position, indirect velocity and acceleration estimation, respectively, and evaluate such a formulation on two benchmark datasets: a warehouse security camera dataset that we collected and annotated and a traffic camera open dataset.
\par Results suggest that our PhyOT can track objects in extreme conditions that the state-of-the-art deep neural networks fail while its performance in general cases does not degrade significantly from that of existing deep learning approaches. Results also suggest that our PhyOT components are generalizable and transferable.
\end{abstract}
\section{Introduction}

Deep neural networks learn intricate structures from high-dimension data such as images \cite{du2021neural}. They significantly outperform previous state-of-the-art methods at computer vision tasks such as image classification and object detection on generic datasets such as ImageNet \cite{imagenet}. These neural networks have millions of parameters, requiring large amounts of labeled training data \cite{rezatofighi2017deepsetnet}. Furthermore, these models often lack robustness to real world operating conditions; occlusion, illumination variation, and background cluttering remain serious challenges for deep models.

\par Object tracking is a more challenging task than image classification and object detection \cite{rosten2005fusing}. It requires both object detection in static images and temporal information analysis to predict trajectories. Humans have an intuitive physics engine that allows them to perform well at tracking objects despite a lack of complete information by leveraging temporal information. Humans intuitively determine the velocity of objects and can use (implicit) physics motion laws to estimate their trajectory.

\par Prior work \cite{Bewley2016_sort} has demonstrated that incorporating physics knowledge (i.e., laws of motion) can improve tracking accuracy over vanilla neural networks, especially in the presence of the aforementioned challenges. By formulating the problem as one of sensor fusion, where the tracking network is interpreted as a sensor, and using existent fusion techniques (e.g., Kalman filtering \cite{Wojke2018deep}), it is possible to improve accuracy, even when prior knowledge is naively simple (e.g., constant velocity, in the case of prior work).

\par Our approach extends this work: by retaining the formulation of the problem as one of sensor fusion, we add an additional ``sensor'' (specifically, a second neural network performing optical flow) that can provide more accurate priors for velocity by determining acceleration at runtime. Our primary hypothesis can thus be stated as:
\begin{itemize}
    \item ``In object tracking, capturing additional information from the underlying physics can improve tracking accuracy, when this additional information is fused with other data appropriately.''
\end{itemize}

We test this hypothesis by extending prior work (\cite{Bewley2016_sort}) with an Optical Flow ``sensor'' that is leveraged to provide indirect velocity estimation, and compare accuracy over two datasets, in relation to prior work, and a vanilla neural network. Specifically, this paper offers the following contributions:

\begin{itemize}
    \item We demonstrate the use of Optical Flow as a side-channel estimator for object velocity in a deep learning framework.
    \item We introduce an acceleration estimator that estimates the acceleration of the object.
    \item We linearize nonlinear trajectories with linear systems driven by the estimated accelerations.
    \item We describe the incorporation of above components in a Kalman filtering system that combines it with an object tracking neural network.
    \item We empirically evaluate tracking accuracy using three different deep learning baseline trackers as sensors, over two benchmark suites.
\end{itemize}

The remainder of this paper is organized as follows. Section \ref{sec:prior_work} briefly reviews the existing work in object tracking in deep learning and heuristic/algorithmic approaches, as well as the intersection of the two. Section \ref{sec:hybrid} describes our hybrid model, formulating its fundamental principles and operation. Section \ref{sec:experiment} presents our experimental setup and results, and Section \ref{sec:conclusions} concludes the paper.

\section{Previous work}\label{sec:prior_work}

\noindent\textbf{Deep learning for visual tracking.} Deep learning approaches ignore the knowledge about object motion and use modern neural network architectures to directly track the target. Two popular paradigms
are Siamese networks \cite{siamfc}\cite{siamrpn}\cite{TransT} and discriminative correlation filter \cite{atom}\cite{dimp}\cite{tomp}. Siamese networks take a pair of images, the search image and the template, as input and find the location of the template in the search image. Each image is applied an identical transformation $\phi$, modeled by a pretrained deep neural networks on object classification task, to get a feature embedding. The similarity measure between two embeddings is modeled or learned via neural networks to locate the template in the search image. While Siamese networks achieve good results in accuracy and speed \cite{siamfc}\cite{goturn}\cite{TransT} they do not incorporate information from the background region or previous tracked frames into the model prediction and they are yet to reach high level of robustness \cite{vot18}. On the other hand, discriminative correlation filters \cite{atom}\cite{dimp}\cite{tomp} learn a target classification module online during prediction. This approach aims to distinguish the tracked object from the background through learned discriminative correlation filters.
While these models are state-of-the-art in object tracking, they do not leverage physics priors of objects motion (if applicable) and lack robustness when there is severe background cluttering present like indoors, e.g. warehouse, and traffic environments that we consider. 

 \noindent\textbf{Model-based tracking.} Kalman filtering \cite{kalman}, a recursive solution to the linear filtering problem for time series data, is one of the most popular solutions to guidance and tracking in control systems. Kalman filters combine the prior prediction using dynamics models and the observation from sensors to make a more accurate estimate of the state. SORT \cite{Bewley2016_sort} and DeepSORT \cite{Wojke2018deep} adopt Kalman filtering into object tracking in videos by performing the filtering over detection outputs of deep neural networks. They employ a simple linear kinetic model with constant heuristic velocity. These methods have shown promising results that provide efficient tracking algorithms in terms of performance and computational time. However, Kalman filtering is originally used in a control system where the control input is known. In object tracking from videos, we do not know the control input (e.g., the power supplied to the moving object that results in its acceleration). The dynamics model in \cite{Bewley2016_sort}\cite{Wojke2018deep} assumes zero control input that translates to constant velocity. This fails when the object is moving nonlinearly.
 
 \noindent\textbf{Optical flow for tracking.}
Optical flow reflects the brightness variation caused by the movement of an object on camera. Even though brightness variation can be due to a change in brightness of a still object, optical flow generally gives indication of movement between two images. Some traditional tracking algorithms use optical flow alone to track objects. Optical flow can also be used to generate good candidate search regions for the target \cite{opticalflow}. However, there is not much work that incorporates optical flow in a deep learning framework.


\section{PhyOT: a Hybrid Tracking model}\label{sec:hybrid}

We define ``hybrid'' models to mean computer vision solutions that combine deep learning techniques with heuristic/algorithmic techniques, towards improved metrics compared to purely deep learning or purely heuristic/algorithmic solutions.

\subsection{Background: the SORT approach}

SORT \cite{Bewley2016_sort} attempts to improve upon pure deep-learning approaches by integrating prior knowledge, in the form of a motion model, in the estimation of object position. Let the actual center position (in pixel coordinates) of the object at time $t$ be

\begin{equation}
    \textbf{L}_t = \begin{bmatrix} p_x \\ p_y \end{bmatrix}_t 
\end{equation}

Let $\widehat{\textbf{L}}_t$ denote the estimation of $\textbf{L}_t$. It is estimated by predicting priors of object position, denoted $\textbf{L}_{t|t-1}$, and refining the estimation as a function of prior and observed positions, denoted $\textbf{L}^{\textrm{obs}}_t$, as per Kalman filtering equations. 

$\textbf{L}_{t|t-1}$ is the prior expectation of position, as given by equation ~\eqref{eq:SORT_prior}

\begin{equation}
    \textbf{L}_{t|t-1} = \widehat{\textbf{L}}_{t-1} + \begin{bmatrix} v_x \\ v_y \end{bmatrix}^{\textrm{est}}_{t-1}
    \label{eq:SORT_prior}
\end{equation}

$\textbf{L}^{\textrm{obs}}_t$ is the observed position at time $t$, i.e., the output of the neural network, replacing the ``sensor'' in the traditional model of Kalman filtering. In the prior work, the authors of SORT demonstrate that this approach outperforms vanilla neural network approaches, even with naive estimation of velocities (in the example, the authors assume zero \textit{y} velocity and constant \textit{x} velocity).
\subsection{Improved velocity estimation}

We make the motion model closer to the real world dynamics by removing the constant velocity assumption. Our approach utilizes a second ``sensor'' that provides us with indirect observation of the tracked object velocities, independent of the tracking network: specifically, we use Optical Flow (OF \cite{lei2009optical}) to approximate the velocity of the tracked object.

Suppose a pixel at location $(x,y)$ at time $t$ with intensity $I(x,y,t)$ moves to a location $(x+\Delta x,y+\Delta y)$ at time $t+\Delta t$, then
\begin{equation}
    I(x,y,t) = I(x+\Delta x,y+\Delta y, t+\Delta t)
\end{equation}

Then the optical flow $V(x,y,t)$ at point $(x,y)$ at time $t$ is defined as $\begin{bmatrix} V_x(x,y) \\ V_y(x,y) \end{bmatrix}$ where $V_x(x,y) = \frac{\Delta x}{\Delta t}$ and $V_y(x,y) = \frac{\Delta y}{\Delta t}$. Assuming the movement to be small and the image brightness to be constant, we can estimate $(V_x,V_y)$ by expanding out the right-hand-side (R.H.S.) with Taylor series truncating the higher order,

\begin{equation}
    \frac{\partial I}{\partial x}V_x+\frac{\partial I}{\partial y}V_y+\frac{\partial I}{\partial t} = 0
\end{equation}

\par Our hypothesis is that an independent velocity observation can improve the overall position estimation, by providing more realistic measures of the actual velocity of the object.

\subsection{Attention mask}
\label{subsection:attention_mask}
In cluttered environments, there are multiple objects moving at the same time or multiple stationary objects in the background. We create an attention mask $M$ based on optical flow and prior knowledge of velocity $v_{t-1}$ to distinguish pixels belonging to the tracked object

\begin{equation}
    M(v_{t-1})[x,y,t] = \mathbbm{1}_{\|V(x,y,t)-v_{t-1}\|_2 \leq g(v_{t-1})}
    \label{eq:attention_mask}
\end{equation}
where $g$ is a function enforcing physical constraints. For example, $g(v) = v\frac{\pi}{6}$ enforces that objects can not turn more than 30 degrees.

\subsection{Acceleration estimation}\label{subsection:acceleration}
In practice, platforms (robots, cars, forklifts) have non-linear dynamics. We capture these by still assuming linear dynamics, like in SORT, but driven by an unknown acceleration that lets the vehicle maneuver (for example, in plane rotations). We add an acceleration module (see Fig\ref{fig:model_architecture}) to estimate the unknown accelerations. Since the acceleration is the derivative of velocity, we use two consecutive optical flows to estimate acceleration. 
 For a pixel $(x,y)$ at time $t$ that moves to $(x+\Delta x,y+\Delta y)$ at time $t+1$, the acceleration of the pixel is estimated by
 
\begin{equation}
    \begin{bmatrix} a_x(x,y) \\ a_y(x,y) \end{bmatrix}_t = \begin{bmatrix} V_x(x+\Delta x,y+\Delta y) \\ V_y(x+\Delta x,y+\Delta y)\end{bmatrix}_{t+1} - \begin{bmatrix} V_x(x,y) \\ V_y(x,y)\end{bmatrix}_{t}
    \label{eq:our_accelerations}
\end{equation}

To align the optical flow $V_{t+1}$ with $V_{t}$, we define a
warping operator $w(I, V)$ that backwarps the image $I$ according to the flow field $V$.

\begin{equation}
    w(I, V)(x,y) = I(x+\Delta x,y+\Delta y)
\end{equation}
 
 We apply the attention mask $M_t$ in~\eqref{eq:attention_mask} to $w(V_{t+1},V_t)$ and $V_{t}$ and use these as inputs to an acceleration estimator, modeled by a neural network, to estimate the acceleration of the tracked object.
\subsection{Kalman integration}
\label{subsection:kalman}
Our approach is based on interpreting each pre-processing stage (outputting position and velocity observations, respectively) that operates on true (physical) sensor output (camera) as individual ``sensors'', such that the problem can be formulated as one of sensor fusion, implemented through Kalman filtering.

\begin{figure*}[htb]
\centering
\includegraphics[width=0.75\textwidth]{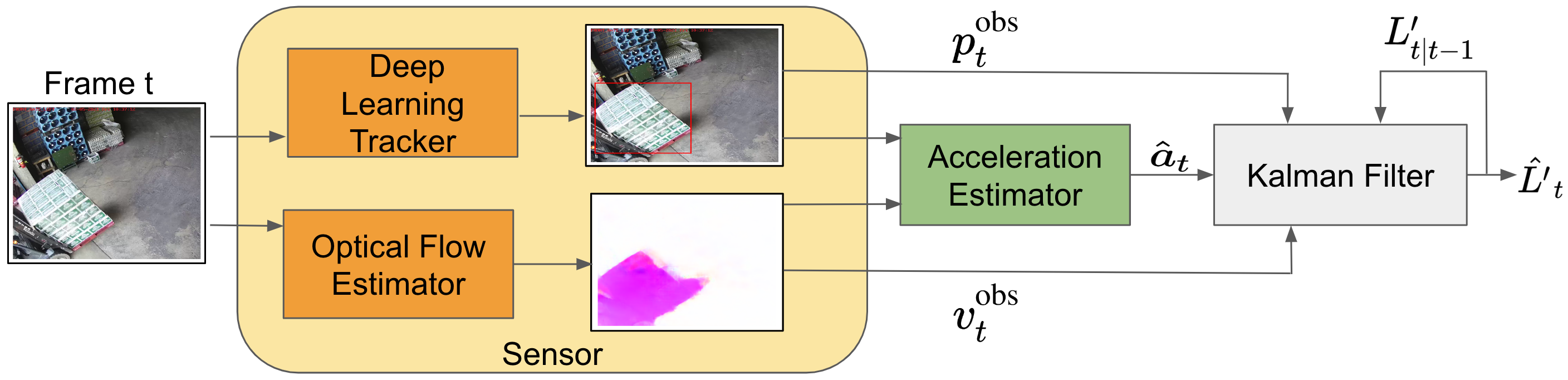}
\caption{PhyOT architecture. The virtual sensor comprises of deep learning tracker and optical flow estimator that separately compute the bounding box and the optical flow, respectively. These outputs are deemed as observations in the Kalman filtering context. The optical flows are used to estimate the acceleration. The Kalman filter takes in observations and estimated acceleration to update the prior state to get a better estimation of the state.}
\label{fig:model_architecture}
\end{figure*}

The full architecture is shown in Figure \ref{fig:model_architecture}. Our solution primarily consists of three components:
\begin{itemize}
    \item \textbf{Sensors.} Virtual ``sensors'' extract physics quantities as transformations from high dimensional inputs (e.g., images). Virtual sensors can be heuristic algorithms or deep neural networks.
    \item \textbf{Kalman filter.} The Kalman filter processes observed physics quantities and estimates the desired information by fusing sensor input, adjusted by prior expectation. In our experiments, sensor input consists of position and velocity estimations given by two neural networks, respectively, and prior expectation is given by Newtonian mechanics.
    \item \textbf{Acceleration Estimator.} The acceleration allows us to linearize nonlinear motions.
\end{itemize}

Let the augmented state (now including velocities) be
\begin{equation}
    \textbf{L}^\prime_t = \begin{bmatrix} p_x \\ p_y \\ v_x \\ v_y \end{bmatrix}_t
\end{equation}

The physics system model formulates the behavior as 
\begin{equation}\label{eq:dynamic sys}
    \begin{split}
    &\textbf{L}^{\prime}_{t} = \textbf{A}\textbf{L}^{\prime}_{t-1} + \textbf{B}\begin{bmatrix} a_x \\ a_y \end{bmatrix}_{t-1} + w_{t-1}, 
        \mbox{ $w_{t-1}\sim \mathcal{N}(0,\textbf{Q})$} \\
        &\textbf{L}^{\prime \textrm{obs}}_t = \textbf{L}^{\prime}_{t} + \nu^{\textrm{obs}}_{t}, \mbox{ $\nu^{\textrm{obs}}_{t-1}\sim \mathcal{N}(0,\textbf{R})$}
    \end{split}
\end{equation}

where
\begin{equation*}
        \textbf{A}=\begin{bmatrix} 1 & 0 & 1 & 0\\ 
        0 & 1 & 0 & 1 \\ 0 & 0 & 1 & 0 \\0 & 0 & 0 & 1 \end{bmatrix},  \textbf{B}=\begin{bmatrix} 0 & 0 \\ 
        0 & 0 \\ 
        1 & 0 \\  
        0 & 1 \end{bmatrix}
\end{equation*}

Our estimation $\widehat{\textbf{L}^\prime}_t$ is given by:

\begin{equation}
    \widehat{\textbf{L}^{\prime}}_t = \textbf{L}^\prime_{t|t-1} + K_t (\textbf{L}^{\prime \textrm{obs}}_t - \textbf{L}^\prime_{t|t-1})
    \label{eq:our_Kalman}
\end{equation}
where $\textbf{L}^\prime_{t|t-1}$ is the prior expectation of position, as given by equation~\eqref{eq:our_prior}

\begin{equation}
    \textbf{L}^\prime_{t|t-1} = \textbf{A}\widehat{\textbf{L}^\prime}_{t-1} + \textbf{B}\begin{bmatrix} a_x \\ a_y \end{bmatrix}_{t-1}
    \label{eq:our_prior}
\end{equation}
, $\textbf{L}^{\prime \textrm{obs}}_t$ is the observed position, i.e., the concatenation of the outputs of the tracking neural network and optical flow.

Unlike in a control system, the actual acceleration $\begin{bmatrix} a_x \\ a_y \end{bmatrix}_t$ is unknown. We use the acceleration estimator scheme described in section \ref{subsection:acceleration} to estimate the acceleration.

\par The Kalman gain $\textbf{K}_t$ is given by the set of equations:

\begin{equation}
    \textbf{K}_t = \boldsymbol{\Pi}_{t|t-1}(\boldsymbol{\Pi}_{t|t-1}\textbf{R})^{-1}
    \label{eq:our_Kalman}
\end{equation} 
\begin{equation}
    \boldsymbol{\Pi}_{t} = (\textbf{I}- \textbf{K}_t)\boldsymbol{\Pi}_{t|t-1}
\end{equation} 
\begin{equation}
    \boldsymbol{\Pi}_{t|t-1} = \textbf{A}\boldsymbol{\Pi}_{t-1}\textbf{A}^T+\textbf{Q}
\end{equation}

\section{Experiments}\label{sec:experiment}

Since we want the motion model to be valid in 2D in videos, we consider video datasets with static cameras (e.g., surveillance cameras) and physics-based objects such as cars. We require the frame of reference to be fixed and the motion of objects to be predictable in order to leverage physics prior knowledge. Extensions of the work to non-static cameras can be done by leveraging frame-of-reference stabilization techniques such as \cite{Selfie21}, but that is beyond the scope of this paper.

We run the experiments on two datasets: Warehouse dataset and CityFlowV2 dataset.
For evaluation metrics, we follow the standard protocol and use the One-Pass Evaluation (OPE) and measure the success plot of different trackers. The trackers are ranked using the Area Under the
Curve (AUC) of their success plots. 

\subsection{Warehouse Experiment}

 \textbf{Warehouse Dataset.} Warehouse Dataset\footnote{The data is available at smartwarehouse2023.cmkl.ac.th \label{warehousedata}} is a dataset collected from CCTV cameras from operating warehouses. The videos contain the activity of inflow and outflow of different products carried by forklifts in the warehouses as well as miscellaneous human activities. Annotators tracked one particular product among different types of products in the warehouses. Challenging conditions in warehouses included illumination variation, background clutters due to existence of distractors, occlusion and nonlinearity of the motion.
In total, we collected 680 sequences from all the videos and split into 140 sequences for training (3871 frames) and 100 sequences (11594 frames) for testing. The training sequences and testing sequences are from different camera positions.
 \\

\noindent\textbf{Baseline.} Our baseline trackers are the state-of-the-art deep learning models for object tracking: TransT \cite{TransT}, DiMP \cite{dimp} and ToMP \cite{tomp}. All models use ResNet50 \cite{resnet} pretrained on ImageNet \cite{imagenet} dataset as backbones. Each model is pretrained with the combined training splits of LaSOT \cite{lasot}, TrackingNet \cite{trackingnet}, COCO \cite{coco}, and GOT10k \cite{got10k} datasets. Since the specific product we are tracking in the warehouse is not a generic object, we fine-tune each baseline tracker with the training set of warehouse data.

\noindent\textbf{Our approach.} We use each baseline tracker as a deep learning tracker component in \ref{fig:model_architecture}. We use Spynet \cite{spynet} as optical flow estimator and the Kalman filter described in \ref{subsection:kalman} .

\noindent\textbf{Training.} We only train the acceleration estimator that takes masked optical flows and estimates the acceleration. The acceleration estimator is modeled using a convolution neural network (CNN) with four convolutional layers followed by two fully-connected layers with ReLU activation. The CNN is optimized using RMSProp with initial learning rate of $0.00001$.

\begin{figure}[htbp]
     \centering
     \begin{subfigure}[b]{0.23\textwidth}
     \centering
\includegraphics[width=\textwidth]{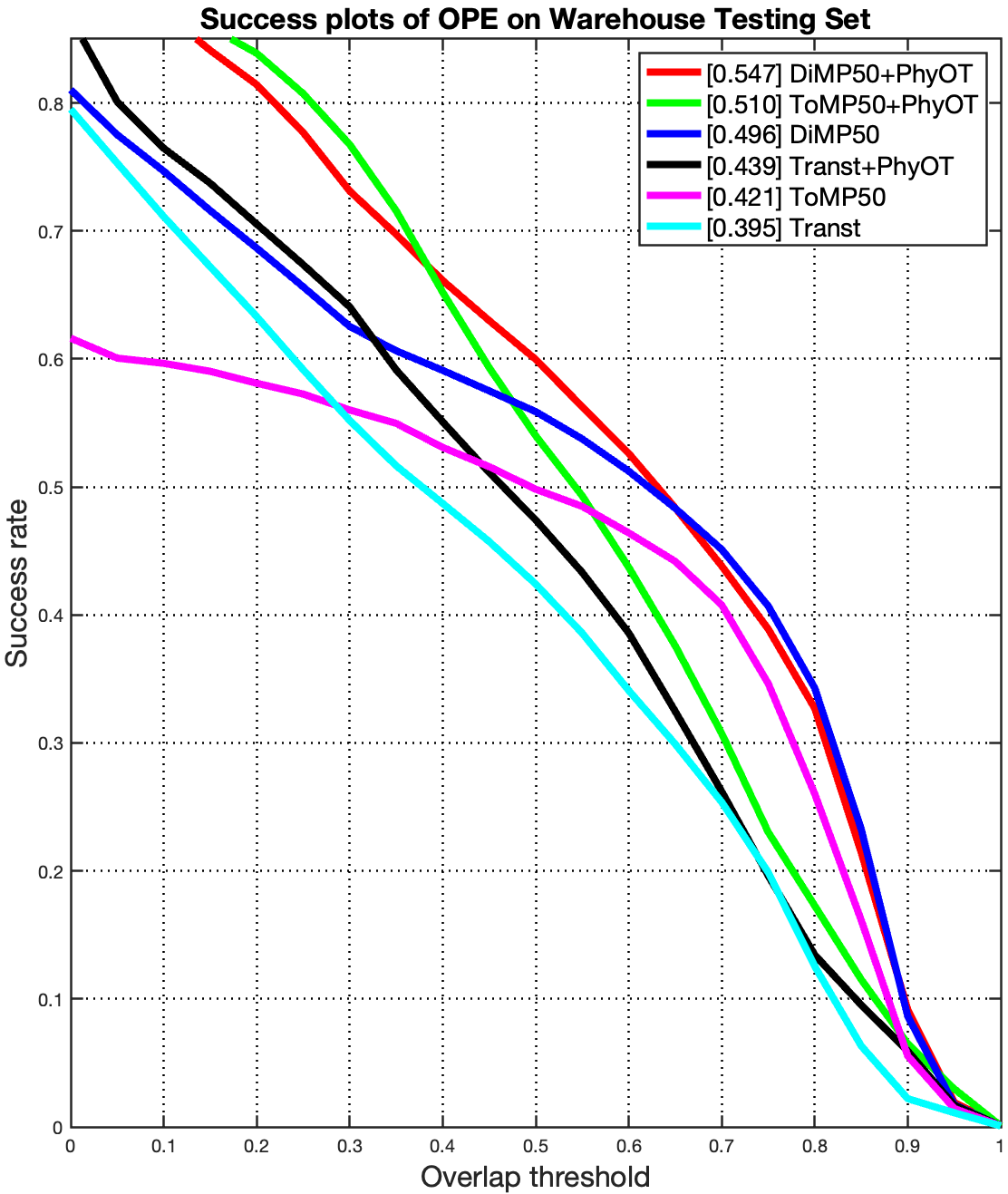}
\caption{Warehouse Test set}
\label{fig:warehouse_success_plot}
     \end{subfigure}
     \begin{subfigure}[b]{0.23\textwidth}
         \centering
\includegraphics[width=\textwidth]{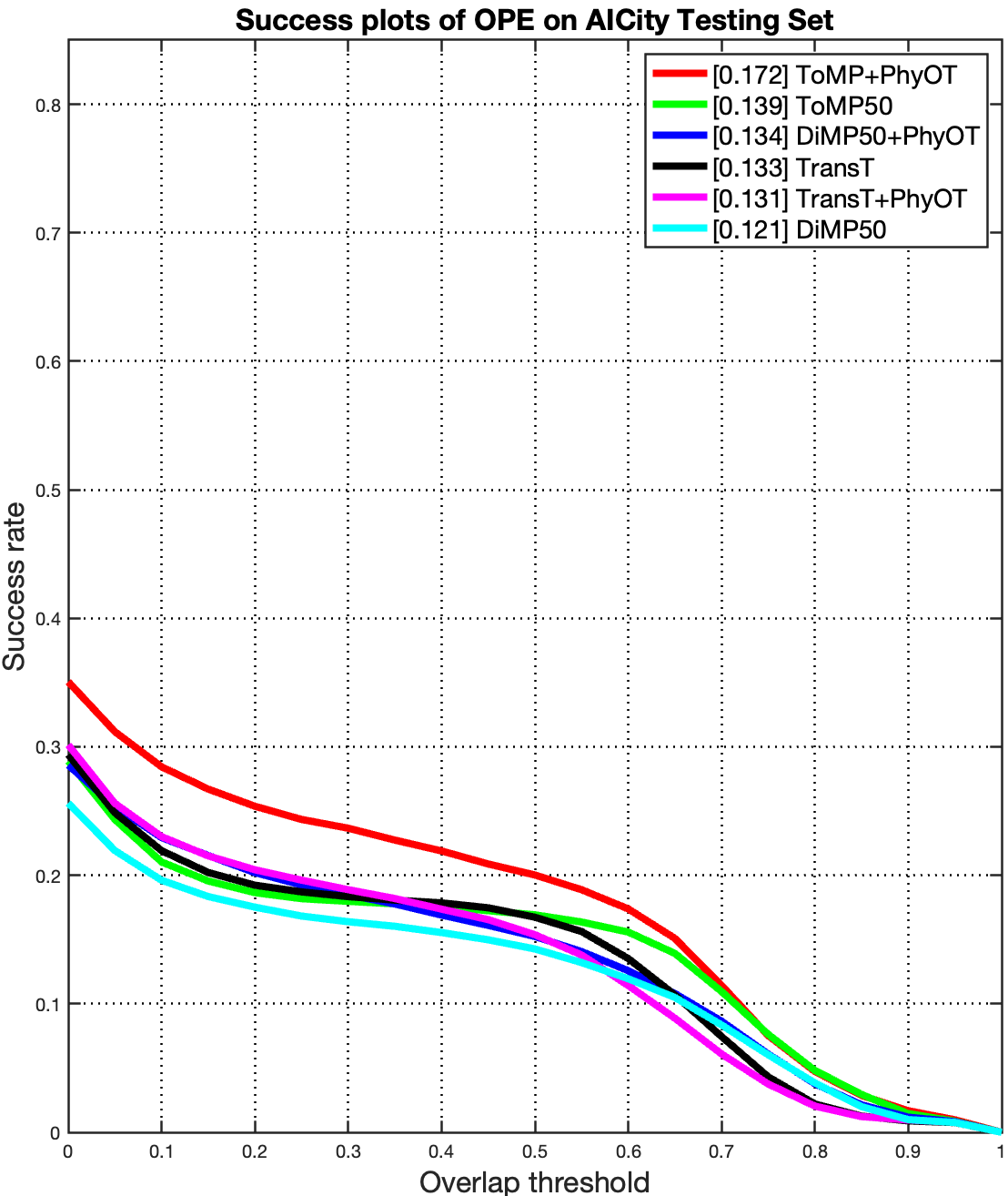}
         \caption{CityFlowV2 Test set}
         \label{fig:aicity_success_plot}
     \end{subfigure}
        \caption{Success plots of OPE on both test sets. AUC is reported in the legend}
        \label{fig:success_plot}
\end{figure}

\noindent\textbf{Experimental results.} The success plot of OPE is shown in Figure \ref{fig:warehouse_success_plot}.

\begin{figure}[htbp]
\centering
\includegraphics[width=0.5\textwidth]{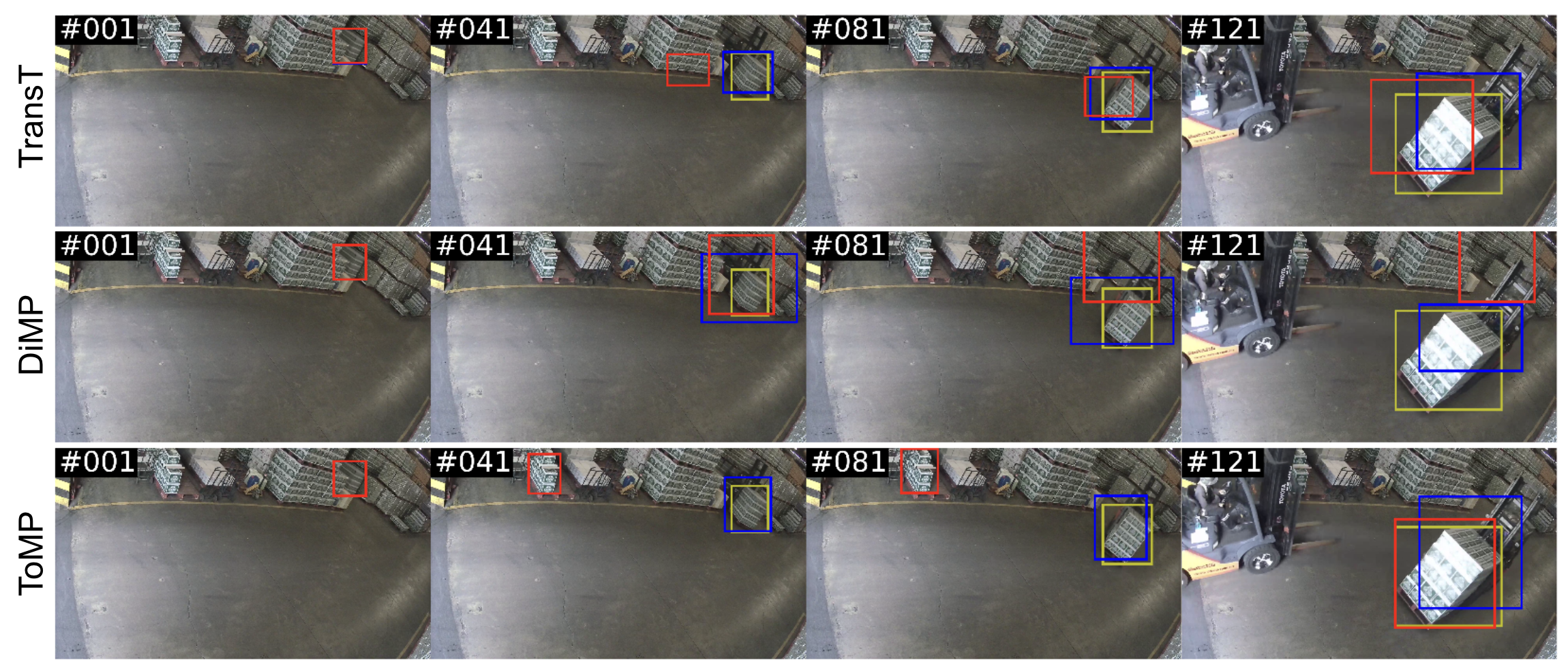}
\caption{Visual comparison between baseline trackers and trackers with PhyOT. Each row shows predicted bounding boxes of the baseline tracker (red), the baseline tracker with PhyOT (blue) and groundtruth (yellow). \textit{From top to bottom}: using TransT, DiMP and ToMP as a baseline tracker, respectively.}
\label{fig:warehouse_qualitative}
\end{figure}

For each deep learning tracker, adding the PhyOT scheme improves their performance by at least $10\%$. Figure \ref{fig:warehouse_qualitative} shows that PhyOT framework can track in the presence of distractors whereas baseline trackers fail.


\subsection{Traffic Experiment}

 \textbf{CityFlowV2 Dataset.} CityFlowV2 \cite{Tang19CityFlow}\cite{Naphade21AIC21} is a real-world data from traffic surveillance cameras collected over 3.25 hours from 40 cameras. The data is split into training set (58.43 minutes), validation set (136.60 minutes) and test set (20.00 minutes). We randomly sample from the train and validation data to test our approach since offline evaluation is not possible on test data.

\noindent\textbf{Baseline.} We use pretrained TransT, DiMP and ToMP as baseline trackers. Since a car is a generic object present in common large-scale dataset, we do not finetune these trackers with CityFlowV2 data.

\noindent\textbf{Training.} Since the acceleration estimator is learning general physics properties (i.e., regressing an acceleration from two optical flows), we argue that it is \textit{class} independent. Hence, we use the trained acceleration estimator 

\noindent\textbf{Experimental results.} The success plot of OPE is shown in Figure \ref{fig:aicity_success_plot}. For each deep learning trackers, adding the PhyOT scheme improves their performance (ToMP and DiMP) or does not degrade the performance significantly (TransT). This demonstrates that our acceleration estimator is generalizable to track other classes of objects that have similar motions.

\section{Conclusions}\label{sec:conclusions}

We present PhyOT, a hybrid model that conceptualizes deep neural networks as ``sensors'' in a Kalman filter setup, where prior knowledge, in the form of Newtonian laws of motion, is used to fuse sensor observations and perform improved estimations. Our experimental evaluation, on a PhyOT instance combining two neural networks, performing position and indirect velocity estimation, shows that prior knowledge can indeed improve vanilla deep learning methods, indicating that hybrid models are a promising research direction.
\par Several open questions remain. The most critical one pertains to the level of detail that prior knowledge should model for a given scenario (e.g., 3D physics), and what are the trade-offs between accuracy, computational cost, and performance, of such level of detail. Examining this research question will undoubtedly lead to advances in computer vision, hopefully bridging the divide between heuristics/algorithms and machine learning approaches.



\bibliographystyle{IEEEbib}
\bibliography{refs}

\end{document}